\documentclass[journal,comsoc]{IEEEtran}
\ifCLASSINFOpdf
\else
\fi
\usepackage{array}
\usepackage{url}
\usepackage{graphicx}
\usepackage{amsmath}
\usepackage{multirow}
\usepackage{makecell}
\usepackage{hyperref}
\usepackage{algorithm}
\usepackage{algpseudocode}
\usepackage{xcolor}
\usepackage{caption}
\begin{document}
%
\title{Multi Teacher Privileged Knowledge Distillation for Multimodal Expression Recognition }
%
%
%

\author{Muhammad~Haseeb~Aslam,~\IEEEmembership{Student Member,~IEEE,}, Marco~Pedersoli,~\IEEEmembership{Member,~IEEE,}
        Alessandro~Lameiras~Koerich,~\IEEEmembership{Member,~IEEE,}
        and~Eric~Granger,~\IEEEmembership{Member,~IEEE}
}

%
%

\markboth{Preprint: Multi Teacher Privileged Knowledge Distillation for
Multimodal Expression Recognition}%
{Shell \MakeLowercase{\textit{Aslam et al.}}: Bare Demo of IEEEtran.cls for IEEE Journals}
%



\maketitle


\begin{abstract}
Human emotion is a complex phenomenon conveyed and perceived through facial expressions, vocal tones, body language, and physiological signals. Multimodal emotion recognition systems can perform well because they can learn complementary and redundant semantic information from diverse sensors. In real-world scenarios, only a subset of the modalities employed for training may be available at test time. Learning privileged information allows a model to exploit data from additional modalities that are only available during training.  
State-of-the-art methods for privileged knowledge distillation (PKD) have been proposed to distill information from a teacher model (that combines different prevalent and privileged modalities)  to a student model (without access to privileged modalities). However, such PKD methods utilize point-to-point matching and do not explicitly capture the relational information in the multimodal space. Recently, methods have been proposed to capture and distill the structural information and outperform point-to-point PKD methods. However, PKD methods based on structural similarity are primarily confined to learning from a single joint teacher representation, which limits their robustness, accuracy, and ability to learn from diverse multimodal sources.
In this paper, a multi-teacher PKD (MT-PKDOT) method with self-distillation is introduced to align diverse teacher representations before distilling them to the student.  MT-PKDOT employs a structural similarity KD mechanism based on a regularized optimal transport (OT) for distillation. An additional constraint is added to the loss function to explicitly align the centroids in the student space. 
The proposed MT-PKDOT method was validated on two challenging affective computing tasks: valence/arousal prediction on the Affwild2 and pain estimation on the Biovid Database.  Results indicate that our proposed method can outperform SOTA PKD methods. It improves the visual-only baseline on Biovid data by 5.5\%. On the Affwild2 dataset, the proposed method improves 3\% and 5\% over the visual-only baseline for valence and arousal respectively. Allowing the student to learn from multiple diverse sources is shown to increase the accuracy and implicitly avoids negative transfer to the student model. 
The code is made publicly available at: \url{https://github.com/haseebaslam95/MT-PKDOT}
\end{abstract}

\begin{IEEEkeywords}
Multimodal Expression Recognition, Multi-Teacher Knowledge Distillation, Privileged Knowledge Distillation, Dimensional Emotion Recognition, Pain Estimation. 
\end{IEEEkeywords}

%

\section{Introduction}
\label{sec:intro}
\IEEEPARstart{E}{motion} recognition (ER) in the wild presents unique challenges in the form of environment variability and domain shift due to demographic diversity, pose variations, and partially or completely missing modalities. This has led to a growing interest  
in multimodal ER (MER) systems aimed at mimicking the process of human-like recognition of emotions \cite{Shon2018}. 
\begin{figure}[!t]
 \centering
  \includegraphics[width=1\linewidth]{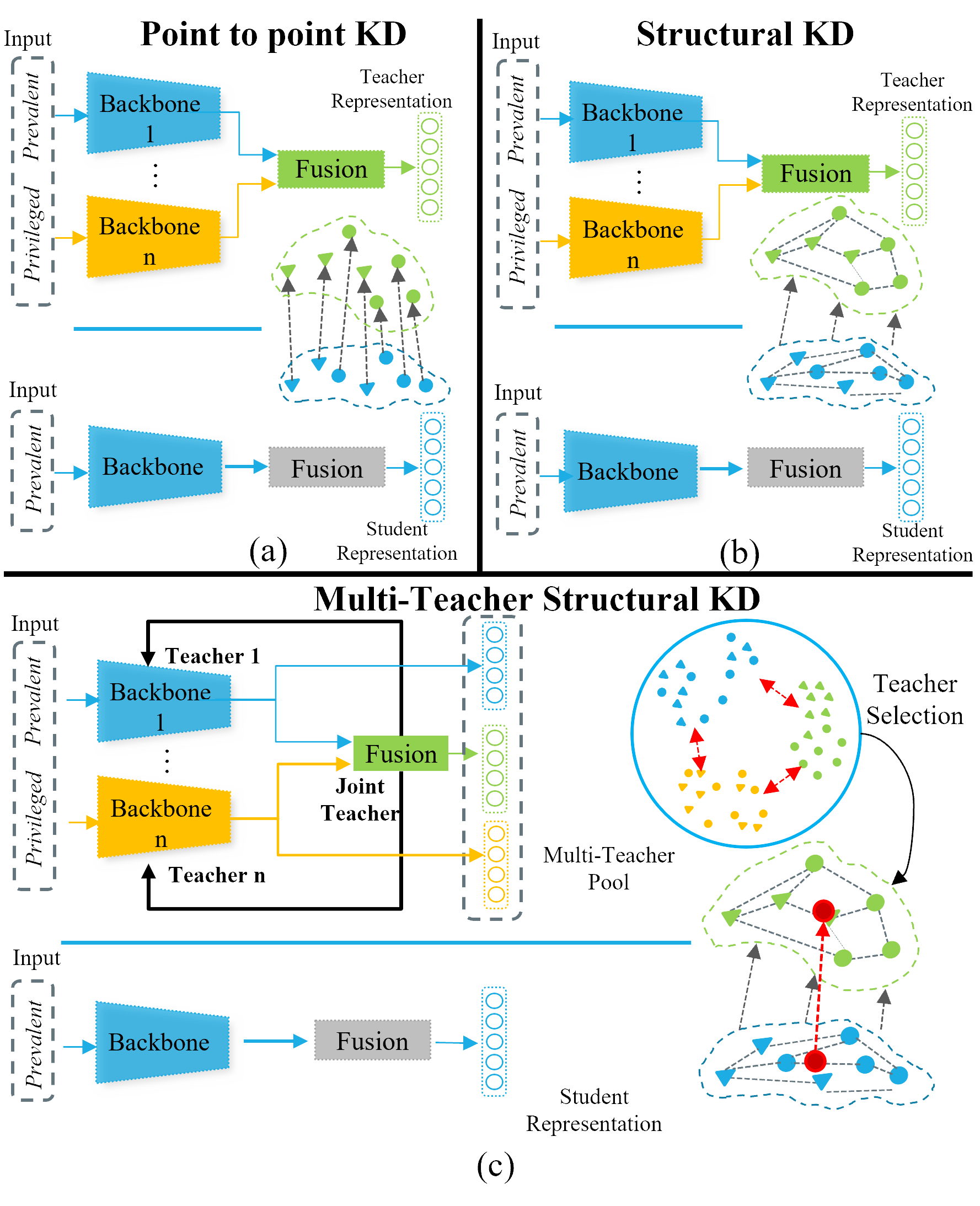}
  \caption{A comparison of PKD methods. (a) The point-to-point based PKD \cite{pkd-aslam} is the vanilla PKD where each point in the student space is matched to the corresponding point in the teacher space. (b) The structural KD-based PKDOT method \cite{aslam2024distilling} captures the relational information and distills it to the student. (c) In contrast, the proposed MT-PKDOT method creates a multi-teacher pool by aligning the backbone teachers with the joint representation through self-distillation and selecting the most confident teacher. A centroid loss is also introduced as an additional constraint to explicitly minimize the $\ell^2$ distance between the centroids of the teacher and student representations.}
  \label{fig:top-mtpkdot}
  \vspace{-15pt}
\end{figure}
MER systems typically outperform their unimodal counterparts due to their ability to capture the redundancy and complementarity information across multiple modalities \cite{TensorFN, Praveen2022AudioVisualFF}. Beyond the additional cost for the capture and fusion of multiple modalities, MER can provide improved accuracy in controlled lab environments because all modalities are available at both train and test times. However, in real-world scenarios, some modalities are difficult or expensive to capture. To overcome this, methods like joint cross-attention \cite{rajasekhar} have been proposed to dynamically assess and weigh the importance of modalities. Even those methods are ineffective where some modalities are entirely missing. 

MER systems employ various modalities including facial, audio, textual, and physiological. Some of these modalities are easier to capture in the wild but signals like electroencephalogram (EEG) \cite{eeg} electrocardiogram (ECG) and electromyography (EMG) \cite{Phan_biovid_phy} are more laborious to be acquired in the wild.    
These physiological signals, in some cases, are more informative than other signals. For instance, the physiological signals have been shown to outperform the visual modality, in tasks like pain estimation \cite{thiam_biovid}. However, in practical contexts, the physiological signals are not always available. Since such systems restrict the subject's mobility and require specialized equipment, most methods will typically rely on prevalent modalities (available at both design time and after deployment), leading to a lower level of system performance. Using this additional information (i.e., privileged modalities) that is only available at the training time may however be leveraged to enhance system performance at test time.

Recently, the learning using privileged information (LUPI) paradigm has been introduced for affective computing \cite{pkd-aslam, aslam2024distilling,pkd-makantasis}. Privileged information (PI) in machine learning (ML) is the information available to the model only at training time but not at inference time \cite{lupi}. For multimodal systems, PI is often the privileged modality that is only available at the train time. LUPI methods have been shown to increase performance for multimodal systems using only the prevalent modalities (available during train and inference) \cite{pkd-makantasis, pkd-aslam, aslam2024distilling, garcia-pi, dai2021learning}. These methods traditionally follow a student-teacher framework for knowledge transfer. An MER model trained with all modalities serves as the teacher network which is comprised of multiple backbones, one specialized per prevalent and privileged modality, and a fusion module to combine their feature representations.  The student is typically a similar model without the PI and its corresponding modules \cite{pkd-aslam}.

Initial studies transferred the PI using the vanilla knowledge distillation (KD) methods, using KL divergence between the softened logits in the teacher and student network \cite{pkd-makantasis}. Researchers have also explored the idea of transferring this knowledge in the feature space using cosine similarity \cite{pkd-aslam} or MSE\cite{pkd-makantasis}. All these methods use point-to-point matching, as shown in Fig. \ref{fig:top-mtpkdot}(a), to distill the knowledge and are, therefore, unable to capture the local structure formed in the teacher space. \textit{Local structure} refers to a more fine-grained knowledge representation, where the distance of each sample from all the other samples in the training batch is calculated. This structural information is formed in the teacher space through the interaction between the prevalent and privileged modality and should be captured to mimic the performance in the student space \cite{aslam2024distilling} as shown in Fig. \ref{fig:top-mtpkdot}(b).  

Zhu et al.~\cite{allenzhu2023-kd-ensemble} explored the concept of distilling knowledge from an ensemble of multiple diverse teachers to a single student. The authors investigated the case where multiple teachers with the same backbone architecture trained on the same dataset and used the same algorithm. The only differentiating element was the random seed initialization. The student was trained jointly using the soft labels learnt by the ensemble as well as the original hard labels.   The authors conclude that combining diverse teacher models each trained using a different random seed introduces improves student accuracy.  Inspired by this work, we explore the idea of multi-teacher KD. In our case, diversity is defined by different modality-specific data and backbone architectures -- each teacher backbone is trained with different modality and fused to form the joint multimodal representation. This fused representation is then aligned with the backbones to provide a diverse multi-teacher pool.


The main contribution of this paper is a multi-teacher privileged KD with optimal transport (MT-PKDOT) method, enabling the student to learn from diverse multimodal sources, thereby enhancing robustness and accuracy, and mitigating negative transfer. Other important contributions are (i) a new loss function that introduces an additional constraint, explicitly forcing the centroids to be aligned simultaneously with the structural similarity-based KD; (ii) a detailed set of experiments performed on two challenging affective computing task problems -- arousal-valance prediction on the Affwild2 and pain estimation on the Biovid heat pain database. Results obtained with a diverse pool of teacher architectures and various modalities show that the proposed MT-PKDOT method is model and modality-agnostic, and can outperform state-of-the-art single-teacher privileged KD methods.

This paper extends our previous work \cite{aslam2024distilling}, where we introduced a PKD method with optimal transport (PKDOT) method using a single fused teacher for additional supervision.
PKDOT introduced a structural KD mechanism to capture and distill the structural dark knowledge. The cosine similarity matrix was used to retain information about the pairwise relationships between all batch samples. An encoder-decoder transformation network (T-Net) was trained during the teacher training stage. Entropy-regularized OT was used to distill the structural dark knowledge. In contrast, this paper extends the single-teacher architecture to a multi-teacher setting. The multiple teachers, i.e., modality-specific teachers and fused teachers, are aligned using self-distillation and modality-specific adapters. The modality adapters are implemented with an encoder-decoder network aimed to project the backbone representations and the joint representation in a common subspace. After the alignment of the modality-specific teachers, the teacher with the minimum error rate is selected. Additionally, the existing  PKDOT method \cite{aslam2024distilling} only relies on the structural similarity loss for the knowledge transfer. In addition to the structural similarity loss employed with PKDOT, the MT-PKDOT calculates the centroids from the selected anchors and minimizes the $\ell^2$ distance to explicitly align the samples. A centroid represents the geometric mean in the teacher and student space. Explicitly aligning these centroids makes the distillation process more controlled as this constraint helps in stabilizing the learning process by reducing the variance in the student model’s parameter updates. The centroid acts as an anchor which reduces oscillations making the distillation process more stable and also helps in faster convergence. Fig. \ref{fig:top-mtpkdot}(b) and (c) compare the PKDOT and MT-PKDOT method.     

The remainder of this paper is structured as follows. Section~\ref{sec:relwork} provides an analysis of state-of-the-art literature related to multimodal MER and PKD.  
Section~\ref{sec:proposed} describes the proposed MT-PKDOT method and its key components. Then, Section~\ref{sec:experiments} provides details on the experimental methodology used for validation (i.e., datasets, evaluation criteria, baselines, and implementation details). Experimental results are presented and analyzed in Section~\ref{sec:results}.  

\section{Related Work}
\label{sec:relwork}

\subsection{Multimodal Emotion Recognition:}

MER seeks to model and understand human emotions by leveraging different input sources like speech, vocal intonations, physiology, text, and facial cues. The seminal work in multimodal deep learning was proposed by Ngiam et al.~\cite{ngiam}, in which both audio and visual modalities were separately encoded. For the fusion, the latent vectors of both modalities were concatenated to form the joint representation. A bimodal deep autoencoder was formed to reconstruct both modalities using separate decoders. 

In the ER, a recurrent network-based fusion technique was proposed by Tzirakis et al.~\cite{tzirakis}, in which the audio features were extracted using a 1D convolutional neural network (CNN), and the visual modality was processed using a ResNet-50 model. The two feature vectors were concatenated and fed to a 2-layer LSTM model to jointly fuse and model temporal dependencies. A joint cross-attention mechanism was proposed by Rajasekhar et al.~\cite{rajasekhar} for overcoming the noisy/missing modality problem, where some modalities are highly unreliable. Separate backbones were trained for both modalities. For the audio modality, discrete Fourier transform (DFT) was applied to the audio signal to obtain spectrograms, which were fed to a ResNet-18. For the visual modality, a 3D CNN was used to extract spatiotemporal features. A joint cross-attention model was used to fuse the two feature vectors. Attention weights were multiplied with the raw features to get the cross-attended features. These cross-attended features overcome the noisy modality problem by dynamically assigning weights.

Pain analysis is another important application in affective computing, with various publicly available datasets. These datasets include induced pain through heat \cite{biovid_ds}, shoulder pain \cite{unbc_mcmaster}, pain estimation in animals \cite{pain_anim} and children \cite{pain_inf} \cite{pain_inf}. Numerous approaches have been proposed for the task of pain detection and estimation. Werner et al.~\cite{werner_2017} developed a pain-specific feature set called facial activity descriptors. A subject-independent deep learning approach was proposed by Dragomir et al.~\cite{dragomir}, in which the concept of residual learning was used for pain estimation from facial images. A data-efficient image-based transformer architecture was proposed by Morabit et al.~\cite{morabit-biovida}. A recurrent net was proposed by Zhi et al.~\cite{zhi-biovid}, where an LSTM model with sparsity was used to overcome the problem of vanishing gradient. A physiological signal-based approach was proposed by Phan et al.~\cite{Phan_biovid_phy} in which ECG and electrodermal activity (EDA) signals were processed using an attention-based method. The LSTM model was used to capture the multi-level context information. For the final prediction, the two modalities were fused at the decision level. Another EDA-based approach was proposed by Lu et al.~\cite{lu_biovid_phy}. Multi-scale EDA signal windows were fed to a residual squeeze and excitation-based CNN. The output of which was combined and fed to a transformer model. Specifically for pain estimation, the physiological signal-based approaches have been shown to outperform the facial image-based approaches. Several multimodal methods have also been proposed to leverage the multiple sources of information. Zhu et al.~\cite{Zhi2021Multimodal_biovid}, Kachele et al.~\cite{Kchele2015BioVid_mm}, and Werner et al.~\cite{wener-multimodal-2014} have proposed methods to combine visual and physiological modalities.

The higher performance and robustness of the multimodal systems come at a cost, usually in terms of time and computational complexity. To minimize the computational complexity, methods like attention bottleneck in transformers \cite{nagrani} have been proposed. However, these methods still fall short in cases where some modalities are entirely missing. On the other hand, the proposed method aims to enhance test-time performance without relying too much on missing or absent modalities.   



\subsection{Knowledge Distillation and Optimal Transport:}

Hinton et al.~\cite{Hinton2015} proposed the seminal work in KD for model compression in a teacher-student setting. The lightweight student model is trained with additional supervision from the accurate yet cumbersome teacher model. The temperature in the teacher's softmax is increased to get less confident yet more informative predictions. This is termed as 'dark knowledge'. These softened predictions are more informative than the one-hot encoding. Romero et al.~\cite{Romero2014} proposed hint learning, essentially distilling from the model's hidden layer instead of softened logits. Since then, several works for feature-based KD have been proposed \cite{Kim2018ParaphrasingCN, heo-ft}. Vanilla KD methods only consider individual samples and solely rely on matching the output activations between teacher and student. A relational KD method was introduced by Park et al.~\cite{Park2019RelationalKD} in which the authors argued that the student model's performance can be significantly improved by distilling the relational knowledge among samples. Inspired from this work, the proposed method also captures the relational information that is formed in the teacher space due to the introduction of the privileged modality.

Optimal Transport (OT) is a well-established mathematical framework for calculating the optimal cost of transforming one probability distribution into another \cite{Villani2008OptimalTO}. OT has seen a rise in ML applications, especially in cases where matching distributions are vital. Other methods for distribution matching, including KL divergence and maximum mean discrepancy, suffer theoretical drawbacks \cite{feydi-iopsd}. KL divergence falls short in cases where two distributions do not overlap, resulting in infinity \cite{kl-db-lohit}. On the other hand, due to its sensitivity to outliers and sample size, MMD does not precisely capture the distance between distributions \cite{mmd_db_gretton}.  Although an expensive solution, OT provides a stable metric for matching distributions. Cuturi et al.~\cite{Cuturi2013SinkhornDL} proposed regularized OT to overcome the computational limitations of OT. Since then, many works have used OT of various applications, including neural architecture search \cite{Yang_2023_CVPR_NAS}, domain adaptation \cite{Luo_2023_CVPR_DA}, model compression \cite{kl-db-lohit}, and, pedestrian detection \cite{Song_2023_CVPR-ped-det}. To the best of our knowledge, this is the first use of OT in a multi-teacher privileged KD setting in the context of expression recognition.


\subsection{Multi-Teacher and Privileged Knowledge Distillation:}

Several multi-teacher KD methods have also been proposed in the literature. Shome et al.~\cite{shome2024emodistill} proposed a multi-teacher speech ER system. Linguistic and prosodic teachers were first trained separately. Speech student was then trained with additional supervision from frozen prosodic and linguistic teachers. Knowledge was transferred at both logit and feature levels. Sarkar et al.~\cite{sarkar2023xkd} introduced a cross-modal KD method for video representation learning. Masked reconstruction, domain alignment, and cross-modal KD steps were performed to enhance the overall performance in video action classification and sound classification. Ma et al.~\cite{Ma_2024_sdt} proposed a method based on self-distillation for ER in conversation, a transformer model was used to learn inter- and intra-modal interactions. A hierarchical gating mechanism was used to dynamically weigh modalities. Soft and hard labels were used to distill knowledge from the multimodal model to each modality. Long et al.~\cite{LONG202312} also proposed a diversified branch fusion for self-KD to effectively utilize the knowledge from the shallow layers via branching. Introducing diverse knowledge from different branches enhances the performance of CNN-based methods.  

The quality and availability of the modalities are a concern in the multimodal ML paradigm. Visual, audio, and, textual modalities can be partially missing due to occlusion, user-initiated muting, and/or transmission/recording errors, etc. For RGB-D data, RGB data is readily available at both train time and deployment, but depth modality is not always available. Similarly for affective computing, physiological signals like EDA, EMG and ECG may only be available at training time but completely missing at deployment. LUPI paradigm can be used to leverage this information at train time only. The concept of PI in ML was proposed by Vapnik and Vashist \cite{lupi}, where additional information, available to the model at training time only, was used to learn more discriminative information, outperforming the traditional ML paradigm of using the same information at training and testing. Many applications, like action recognition \cite{prid-pi} and person re-identification \cite{pi-shang}, have since utilized the concept of LUPI to improve performance or increase robustness. An online action detection privileged KD mechanism was proposed by Zhao et al.~\cite{zhao-pi}, where the PI was the future frames from the video, which were only available to the model during train time. However, in real-life deployment, only historical frames are available. To minimize the student-teacher gap, KL-divergence was used to only update partially hidden features in the student model. 

As the focus in affective computing shifts to more in-the-wild scenarios, privileged KD methods have become more popular due to their ability to perform well in extreme cases of completely missing modalities at test time. Existing methods primarily use  PI as an additional modality to train a superior multimodal teacher, which is then used to distill the information to a student model that does not have access to the privileged modality. A point-to-point matching-based method was proposed by Aslam et al.~\cite{pkd-aslam}. Cosine loss was added to the task loss for the knowledge transfer between the multimodal teacher and the student model. To mitigate the negative transfer from the teacher method, adaptive weighting among the task and KD was used. Makantasis et al.~\cite{pkd-makantasis} proposed a privileged KD method for two tasks; the categorical expression recognition problem used the KL divergence loss, whereas the arousal-valence prediction model was trained using the MSE loss function. Liu et al.~\cite{Liu-pkd} also proposed a privileged-KD-based method in the physiological signals domain, where a GSR-based student model was trained with additional supervision from the teacher model that was trained using EEG and GSR. The KD loss used was the KL divergence. Although these methods improve the performance of the student model, they lack any defined mechanism to capture the relational knowledge in the multimodal teacher space. Capturing the local structures formed in the teacher network space by the introduction of privileged modalities should be distilled to the student network to enhance performance. Furthermore, all of aforementioned PKD methods are confined to learning from a single joint teacher which drastically limits their robustness and the ability to learn from multiple diverse sources. 

In contrast, our proposed method performs self-distillation using the joint multimodal feature space to the modality-specific feature representations. Our goal is to align the teacher representations and effectively create a diverse multi-teacher pool for the students to learn from.

\section{Proposed Approach}
\label{sec:proposed}

\begin{figure*}[!t]
  \centering
   \includegraphics[width=0.9\linewidth]{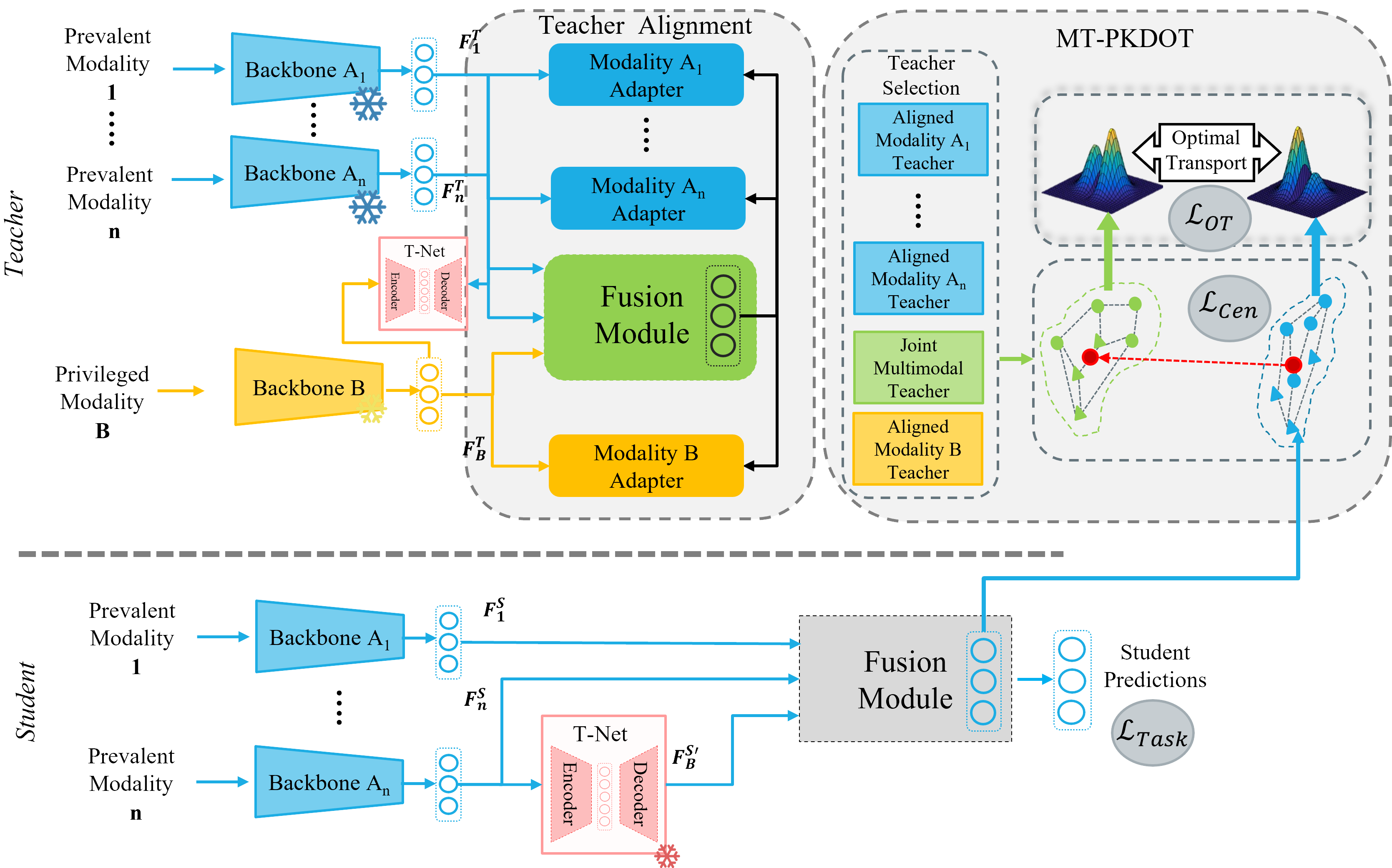}
   \caption{Illustration of the proposed MT-PKDOT method to train the student model. In the multi-teacher pool, the representation of $n$ modality-specific teachers is aligned using self-distillation. Following the selection of the most-confident teacher for the batch, the relational knowledge is captured using cosine similarity matrices. For similarity structure knowledge transfer, entropy-regularized OT is used to match the teacher and student distributions. A centroid loss is also used as an additional constraint to explicitly minimize the distance between the teacher and student.}
   \label{fig:pkdot_main}
   \vspace{-10pt}
\end{figure*}

The MT-PKDOT method relies on the diversity among teacher models to enhance the performance of the student network. In the teacher space, the modality-specific features are aligned with the fused feature vector using modality adapters. The aligned teachers then serve as the additional supervision for the student. At student training, the most confident teacher is selected based on the task performance metric. Further, structural similarity matrices calculation and anchor selection are done similarly to PKDOT \cite{aslam2024distilling}. For the distillation, in addition to the OT loss, a constraint to minimize the centroid loss is also added to simultaneously align the centroid and distill the structural information. 
Fig. \ref{fig:pkdot_main} shows the proposed different modules of the proposed MT-PKDOT method. The remaining subsection presents additional details on the key components of this framework.

\subsection{Teacher Alignment using Self Distillation and Selection:}

Let $\mathbf{F}_i = [\mathbf{F}_{A_1}, \mathbf{F}_{A_2}, \ldots, \mathbf{F}_{A_n}, \mathbf{F}_B]$, where $\mathbf{F}_{A_1}, \mathbf{F}_{A_2}, \ldots, \mathbf{F}_{A_n}$ represent the feature vectors from the first to the $n$-th backbone, and $\mathbf{F}_B$ represents privileged modality feature vector. The joint representation from the fusion network is represented as $\mathbf{F}_{\text{Joint}}$. 
To add diversity to the teacher space, modality-specific representations are also included in the teacher pool. However, the backbone and the joint representations are in separate spaces. To project them in a single space and align them while maintaining diversity, the joint representation and the modality-specific backbone representations are aligned using modality adapters. A modality adapter is an encoder-decoder-based network that takes in the modality-specific features and tries to reconstruct the joint representation, effectively aligning the backbone representations with the joint representation. We use the cosine loss function:
\begin{equation}
 \mathcal{L}_\text{align} = \displaystyle\sum\limits_{i=1}^{n}  1 - \frac{\mathbf{F}_{i} \cdot \mathbf{F}_{\text{Joint}}}{\left \| \mathbf{F}_i \right \| \cdot \left \| \mathbf{F}_{\text{Joint}} \right \|}
 \label{eq:l_cos}
\end{equation}
\noindent where $\mathbf{F}_{i}$ is the feature representation of the $i$-th backbone. 


\subsection{Relational Knowledge Capture:}
Let $\Theta$ be a matrix of dimension $b\times m$ obtained from the teacher network, where $b$ is the batch size and $m$ is the dimension of the feature vector, and $\Theta^\top$ be its transpose of dimension $m\times b$, and $\Phi$ be a matrix of dimension $b\times m$ obtained from the student network, and $\Phi^\top$ be its transpose of dimension $m\times b$.
\begin{equation} \label{calc_cst}
 C_\Theta = \Theta \cdot \Theta^\top
\text{,}\quad
 C_\Phi = \Phi \cdot \Phi^\top
\end{equation}
\noindent where $C_\Theta$ ($C_\Phi$) is a matrix of size $b\times b$ and each element $C_{ij}$ represents the dot product of the $i^{th}$ row of $\Theta$ ($\Phi$) and the $j^{th}$ column of $\Theta^\top$ ($\Phi^\top$)  and can be denoted as:
\begin{equation} \label{calc_elementct}
C_{ij}^\Theta = \sum_{k}^{} \Theta_{ik}.\Theta^{\top}_{kj}
\text{,}\quad
C_{ij}^\Phi = \sum_{k}^{} \Phi_{ik}.\Phi^{\top}_{kj}
\end{equation}

To ensure that the similarity matrices are not heavily influenced by the magnitude, we normalize $C_\Theta$ and $C_\Phi$ as: 
\begin{equation} \label{norm_t}
N_{\Theta} = \sqrt{\sum_{k=1}^n (\Theta_{ik})^{2}}
\text{,}\quad
N_{\Theta}^ \top = \sqrt{\sum_{k=1}^n (\Theta^\top_{ik})^{2}}
\end{equation}
\noindent where $N_\Theta$ ($N_\Phi$)  denotes the normalization vector of size $b\times 1$, computed by calculating the $\ell^2$-norm for each row. $N_\Theta$ ($N_\Phi$) allows us to normalize the values so that the length of the vectors does not dominate the cosine similarity measure. $N_{\Theta}^\top$ ($N_{\Phi}^\top$) is also calculated using Eq.~\eqref{norm_t}, with the exception that it is of dimension $1\times b$.


$S_\Theta$ and $S_\Phi$ are the final similarity matrices of teacher and student networks respectively and are obtained by element-wise division of $C_\Theta$ with the dot product of $N{_\Theta}$ and $N_\Theta^ \top$ and $C_\Phi$ by the dot product of $N_\Phi$ and $N_{\Phi}^\top$ as follows: 
\begin{equation} \label{sim_mat}
S_\Theta = \frac {C_\Theta} {N_\Theta \cdot N_{\Theta}^\top}
 \text{,} \quad
S_\Phi = \frac {C_\Phi} {N_\Phi \cdot N_{\Phi}^\top }
\end{equation}

\subsection{Entropy Regularized Optimal Transport:}
After the teacher ($S_\theta$)  and student ($S_\phi$) cosine similarity matrices are obtained for a given batch. Top-k most dissimilar samples are selected in the teacher space and a new similarity matrix of dimension $b\times k$ is developed, where K represents the number of anchors.  The same indexes are also selected in the student similarity matrix. The \textit{structural dark knowledge} is distilled from the teacher to the student using entropy-regularized OT, defined by:

\begin{equation}\label{eq:ot}
\mathcal{L}_{\text{OT}}(\mu ,\nu) = \int_{\chi \times \chi}^{} \mathcal{O}(S_{\Theta i},S_{\Phi i}) d\pi(S_{\Theta i},S_{\Phi i}) + \epsilon H(\pi)
\end{equation}

\noindent where $S_{\Theta i}$ and $S_{\Phi i}$ refer to each row in the teacher and student similarity matrices, respectively, semantically representing the local structures, $\mu$ and $\nu$ are the marginal distributions. $\mathcal{O}$ and $d\pi$ represent the cost matrix and the transport plan, respectively. $\epsilon$ $>$ 0 is a coefficient and $H(\pi)$ is the entropic regularization defined as:
\begin{equation} \label{eq:entropic_reg}
    H(\pi) = \log \left( \frac{d \pi}{d\mu \cdot d\nu} (S_{\Phi i},S_{\Theta i}) \right)
\end{equation}
${\mathcal{L}_{\text{OT}}}$ is an approximation of the Wasserstein distance between the teacher and student similarity matrices. Minimizing ${\mathcal{L}_{\text{OT}}}$ effectively transfers the relational information to the student.

The task loss ($\mathcal{L}_{\text{Task}}$), either the concordance correlation coefficient (CCC) for regression or the categorical cross-entropy for classification, is calculated with the ground truth ($y$) according to:
\begin{equation} \label{eq:task_loss_ccc}
\mathcal{L}_{\text{Task}}=\left\{
                \begin{array}{ll}
                  1 - \displaystyle\frac{2. \rho \cdot \sigma_{x} \cdot \sigma_{y}} {\sigma^{2}_{x} + \sigma^{2}_{y} + (\mu_{x}-\mu_{y})^{2}} & \text{if regression}\\
                  \\
                  -\displaystyle\sum_{j=1}^N \displaystyle\sum_{i=1}^C y_{i,j}\log(p_{i,j}) & \text{if classification}
                \end{array}
              \right.
\end{equation}
\noindent where $\rho$ is the Pearson correlation coefficient, $\sigma_x$, and $\sigma_y$, are the standard deviations of predicted and ground truth values. $\sigma^2$ shows the variance and $\mu$ is the mean value. $N$ represents the number of samples and $C$ the number of classes. $y_{i,j}$ are the ground truth values and $p'_{i,j}$ are the predicted values.

\begin{algorithm}[t]
\caption{Teacher Alignment with Self Distillation }
\begin{algorithmic}[1]
\State \textbf{Input:} Feature vectors from $n$ backbones  $\mathbf{F}_i = [\mathbf{F}_{A_1}, \mathbf{F}_{A_2}, \ldots, \mathbf{F}_{A_n}, \mathbf{F}_B$ ] joint representation $\mathbf{F}_{\text{Joint}}$ from the fusion network
\State \textbf{Output:} Aligned representations $\mathbf{F}'_{A_1}$, $\mathbf{F}'_{A_2}$ \ldots $\mathbf{F}'_{A_n}$, $\mathbf{F}'_{B}$ 

\State Set learning rate $\alpha$

\For{each training iteration}

    \State Compute cosine loss:
    \State \quad $\mathcal{L}_{\text{align}} = \displaystyle\sum\limits_{i=1}^{n} (1 - \frac{\mathbf{F}'_i \cdot \mathbf{F}_{\text{Joint}}}{\|\mathbf{F}'_i\| \|\mathbf{F}_{\text{Joint}}\|}$)
    
    \State Update representations using gradient descent:
    \State \quad $\mathbf{F}'_{A_1} \gets \mathbf{F}'_{A_1} - \alpha \cdot \frac{\partial \mathcal{L}_{\text{Cos}}}{\partial \mathbf{F}'_{A_1}}$
    \State \quad $\mathbf{F}'_{A_n} \gets \mathbf{F}'_{A_n} - \alpha \cdot \frac{\partial \mathcal{L}_{\text{Cos}}}{\partial \mathbf{F}'_{A_n}}$
    \State \quad $\mathbf{F}'_{B} \gets \mathbf{F}'_{B} - \alpha \cdot \frac{\partial \mathcal{L}_{\text{Cos}}}{\partial \mathbf{F}'_{B}}$

\EndFor

\State \textbf{return} $\mathbf{F'}_1, \mathbf{F'}_2, \ldots, \mathbf{F'}_n$
\end{algorithmic}
\end{algorithm}

\subsection{Multi-Teacher Knowledge Distillation:}

\paragraph{Centroid Alignment }
After the anchor selection in the teacher and student space, in addition to transferring the structural information, we also calculate the centroid in the teacher and student space. This enables us to explicitly minimize the $\ell^2$ distance between the two representations instead of implicitly relying on the structural knowledge transfer module to minimize this distance:
\begin{equation}
\mathcal{L}_{\text{Cen}} = \bigg\Vert \frac{1}{n}\sum_{i=1}^{n} \mathbf{F}_{\text{Teacher}} - \frac{1}{n}\sum_{i=1}^{n} \mathbf{F}_{\text{Student}} \bigg\Vert^2
\end{equation}

\paragraph{Teacher Selection }
This module selects the most confident frozen teacher in each batch, the one with the least task loss, and distills only from that. The student network is then trained using: 
\begin{equation} \label{eq:total_loss}
\mathcal{L}_{\text{Student}} = \alpha \mathcal{L}_{\text{Task}} + \beta {\mathcal{L}_{\text{OT}}} + \gamma\mathcal{L}_{\text{Cen}}
\end{equation}
\noindent where, $\mathcal{L}_{\text{Task}}$, $\mathcal{L}_{\text{OT}}$, and $\mathcal{L}_{\text{Cen}}$ are the task, OT, and centroid loss respectively, and $\alpha$, $\beta$ and $\gamma$ are the weight coefficients.

\paragraph{Mitigation of Negative Transfer. } The proposed method implicitly mitigates the negative transfer. In the teacher selection phase, the proposed method selects the most confident teacher based on the loss calculated from each aligned teacher representation separately. If the loss value from the aligned backbone teacher network is greater than the existing joint teacher loss, the model will resort back to learning from the joint teacher.

\begin{algorithm}
\caption{Teacher Selection and Student Training with Optimal Transport and Centroid Loss}
\begin{algorithmic}[1]
\State \textbf{Input:} Aligned representations of n backbones $\mathbf{F'}_i = \{\mathbf{F'}_1, \mathbf{F'}_2, \ldots, \mathbf{F'}_n\}$ joint representation $\mathbf{F}_{\text{Joint}}$ from the fusion network and student representation $\mathbf{F}_{\text{Student}}$,
\State \textbf{Output:} Updated student representation $\mathbf{F}'_{\text{Student}}$

\State Set learning rates $\alpha, \beta, \gamma$

\For{each training iteration}
    \State \textit{Compute task losses per teacher}:
    

    \State \textit{Teacher selection}:
    \State \quad $\mathbf{F}_{\text{Teacher}} \gets 
    \arg\min\{\mathcal{L}_{\text{Task}}^{\text{Teacher}_1}, \ldots, \mathcal{L}_{\text{Task}}^{\text{Teacher}_n}, 
    \mathcal{L}_{\text{Task}}^{\text{Joint}}\}$
    
    \State \textit{Compute similarity matrices}:
    \State \quad $\mathbf{S}_{\theta} = \mathbf{F}_{\theta} \mathbf{F}_{\theta}^{\top}$
    
    \State \quad $\mathbf{S}_{\phi} = \mathbf{F}_{\phi} \mathbf{F}_{\phi}^{\top}$
    
    \State \textit{Compute OT loss}:
    \State \quad $\mathcal{L}_{\text{OT}} = \text{OptimalTransport}(\mathbf{S}_{\theta}, \mathbf{S}_{\phi})$
    
    \State \textit{Compute Centroids}:
    \State \quad $\mathbf{\overline{G}}_{\text{Teacher}} = \frac{1}{n} \sum_{i=1}^{n} \mathbf{F}_{\text{Teacher}}$
    \State \quad $\mathbf{\overline{G}}_{\text{Student}} = \frac{1}{n} \sum_{i=1}^{n} \mathbf{F}'_{\text{Student}}$
    
    \State \textit{Compute Centroid loss}:
    \State \quad $\mathcal{L}_{\text{Cen}} = \|\mathbf{\overline{G}}_{\text{Teacher}} - \mathbf{\overline{G}}_{\text{Student}}\|^2$
    
    \State \textit{Compute student task loss}:
    \State \quad $\mathcal{L}_{\text{Task}}^{\text{Student}} = \mathcal{L}_{\text{Task}}(\mathbf{F}'_{\text{Student}})$
    
    \State \textit{Compute total loss}:
    \State \quad $\mathcal{L}_{\text{Student}} = \alpha \cdot \mathcal{L}_{\text{Task}}^{\text{Student}} + \beta \cdot \mathcal{L}_{\text{OT}} + \gamma \cdot \mathcal{L}_{\text{Cen}}$
    
    \State \textit{Update student weights}:
    \State \quad $\mathbf{F}'_{\text{Student}} \gets \mathbf{F}'_{\text{Student}} - \eta \cdot \frac{\partial \mathcal{L}_{\text{Student}}}{\partial \mathbf{F}'_{\text{Student}}}$
\EndFor

\State \textbf{return} $\mathbf{F}'_{\text{Student}}$
\end{algorithmic}
\end{algorithm}

\section{Experimental Methodology}
\label{sec:experiments}
 \begin{figure*}[!t]
  \centering
   \includegraphics[width=0.9\linewidth]{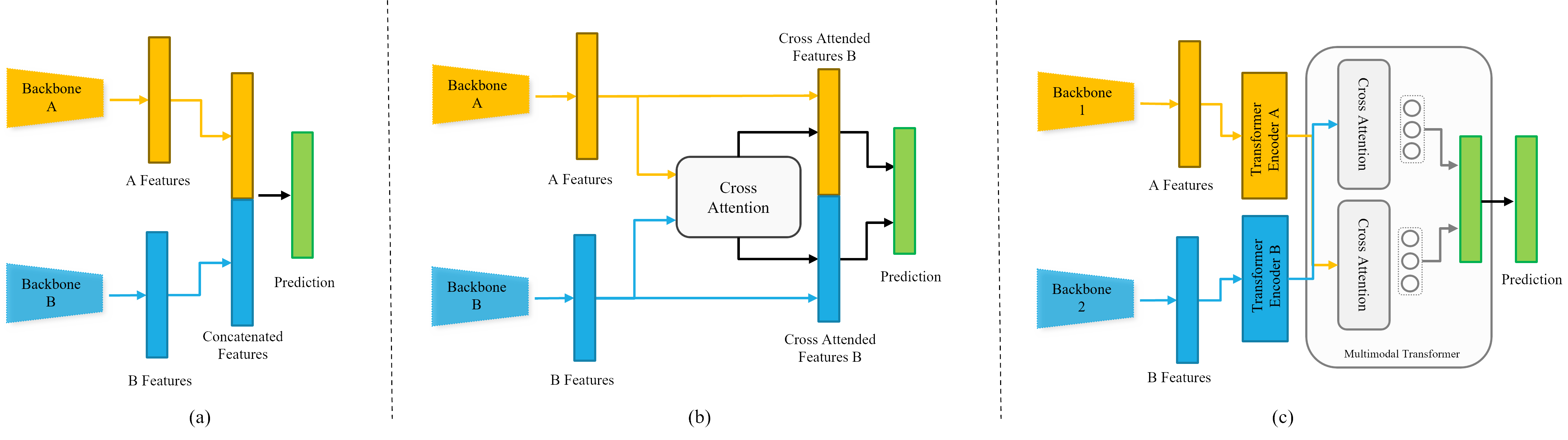}
   \caption{Illustration of the various fusion architectures employed to obtain the fused representation in the teacher space: (a) feature concatenation, (b) joint cross attention \cite{rajasekhar}, (c) multimodal transformer \cite{waligora2024joint}}
   \label{fig:teacher_arch}
   \vspace{-10pt}
\end{figure*}

\subsection{Datasets and Evaluation Protocol}
\noindent \textbf{1) Biovid Heat Pain Database} is among the most popular databases for pain estimation. The dataset is divided into 5 parts. We use part A and part B of the dataset. Part A of the dataset contains videos and the physiological signals including GSR, EMG and ECG. The dataset is annotated for discrete labels for pain intensities, where BL refers to 'baseline/no pain' and PA1-PA4 refers to increasing pain intensities. The Biovid part A has a total of 87 subjects, with each subject having 100 videos, corresponding to 20 videos per class, which results in a total of 8700 videos. 20 subjects did not exhibit any noticeable response to the pain stimulus, so some studies report results for the remaining 67 subjects. For the validation of the proposed method, we report results on the entire dataset of 87 subjects. In addition to the modalities available in part A, part B of the dataset has facial EMG as well and is recorded for a total number of 86 subjects, corresponding to 8600 videos. The Biovid dataset does not come with predefined training, validation, or test splits, so many studies have validated their methods using cross-validation. Following that, we also validate the proposed method using 5-fold cross-validation. The performance metric used is accuracy.

\noindent \textbf{2) Affwild2 } is among one of the most comprehensive datasets for in-the-wild ER \cite{affwild}. The dataset consists of a total of 564 videos with variable lengths. The dataset comes with annotations for three main affective computing tasks, i.e., the action unit detection problem, categorical expression recognition, and the continuous arousal/valence prediction problem. For the validation of the proposed method, the arousal/valence prediction set of the Affwild2 dataset is used. Affwild2 contains videos that are highly diverse in terms of gender, age, ethnicity, and capture conditions, etc. This makes Affwild2 a challenging dataset in terms of generalization. The dataset comes with a pre-defined split for the training, validation, and test subsets with 351, 71, and 152 videos, respectively. The performance metric used for evaluation is CCC. The test annotations for the dataset are not publicly available for running the ABAW Challenge. Consequently, many studies report their results on the validation set. 

\subsection{Implementation Details}
\subsubsection{\textit{Biovid Dataset}}

\textbf{Teacher} For the Biovid (B) dataset, the EMG signals are first converted in spectrograms of dimension 67$\times$127. The physiological feature extraction is done using a ResNet-18 model with the output feature vector of 512d. For the visual modality, an R3D model is used for feature extraction. The visual backbone takes facial frames of size 112$\times$112 as input and jointly models the spatiotemporal dependencies in the visual modality. For the fusion of the two modalities, a transformer-based fusion model is used (Fig.~\ref{fig:teacher_arch}c) \cite{waligora2024joint}. The visual backbone is optimized using the Adam optimizer with a batch is of 64 and the learning rate of 10$^{-3}$. The two backbone feature vectors are fed to two separate transformer encoders and then to the multimodal transformer. The Query vector is generated from one modality, and the Key and Value vectors are obtained from another modality. The cross-modal transformer outputs the cross-attended features, which are then gated using learnable weights. For the final predictions, the gated cross-attended features are passed to fully connected layers. The fusion module uses the Adam optimizer and a learning rate of 10$^{-4}$ with a batch size of 64.

For the Biovid (A) dataset, the proposed method is evaluated with a relatively simpler feature concatenation-based fusion approach. (Fig.~\ref{fig:teacher_arch}a). The visual backbone is the same as Part B. For the physiological modality, EDA signals are fed to a 1D CNN for feature extraction. The output feature vectors of both backbones are concatenated and fed to an MLP for final classification. The fusion module is optimized using Adam optimizer with a batch size of 64 and learning rate of 10$^{-4}$ 
\textbf{Modality Adapters:} The modality adapters are an encoder-decoder based networks that are used for self-distillation, knowledge from the joint representation is distilled to earlier layers. The modality adapters take in the modality-specific feature representations and align them with the joint feature representation. The network is optimized during the teacher training step using the Adam optimizer and with the learning rate of 10$^{-3}$. 
\textbf{Student:} In the student training step, since the physiological modality is the privileged modality, and the model does not have access to the physiological input source, the backbone is dropped altogether. The visual backbone and multimodal transformer model in the student network are the same as the teacher. In order to hallucinate the privileged modality, T-Net is added. The student is also optimized using Adam optimizer with 128 batch size and a learning rate of 10$^{-4}$. 

\subsubsection{\textit{Affwild2 Dataset}} \label{subsec:affwild}

\textbf{Teacher:} 
For the visual modality, a 3D CNN is used for spatiotemporal modeling. The model takes in cropped and aligned facial images of size 112$\times$112. The model is trained with a batch size of 8 and a learning rate of 10$^{-3}$. For the audio modality, the extracted audio is divided into multiple short segments corresponding to 256 frames in the visual modality. DFT is applied on the audio segments to obtain the spectrograms of resolution 64$\times$107. A ResNet-18, trained from scratch, takes in these spectrograms for the extraction of audio features.  The batch size of 8 is used for the audio modality as well, and the network is optimized using a learning rate of 10$^{-3}$. The two modalities are fused using the joint cross-attention model (Fig.~\ref{fig:teacher_arch}b) ~\cite{praveen-tbbi}. The output feature vectors of the two backbone networks are concatenated and fed to the cross-attention module. The fusion module is optimized using the Adam optimizer with a learning rate of 10$^{-3}$. The batch size for the fusion module is set to 64. 
\textbf{Modality Adapters:} are trained during the fusion training step so the batch size is kept the same as the fusion module. The learning rate is set to 10$^{-3}$, and the network is optimized using the Adam optimizer. 
\textbf{Student:} For the Affwild2 dataset as well, the student network follows roughly the same implementation method as the teacher network. Since the privileged modality is not available to the student, the audio backbone is dropped, and the T-Net is added. To hallucinate the privileged modality features for the student network, the prevalent modality features are fed to the frozen T-Net. The batch size is set to 128 for the student model. Adam optimizer with a learning rate of 5$^{-3}$ is used for the student training.

\section{Results and Discussion}
\label{sec:results}
\subsection{Comparison with the State-of-the-Art}
The proposed method is validated on a variety of fusion architectures and modalities from two different datasets. Further, we also distill in different settings like 'Stronger Enhancing Weaker (SEW)' and 'Weaker Enhancing Stronger (WES). For a more comprehensive analysis, we distill not only the visual modality but also the physiological modality. Table~\ref{tab:summ-mod} summarizes the information on the datasets, the modalities used, the distillation setting, and the target student modality.

\begin{table}[]
\addtolength{\tabcolsep}{-0.2em}
\begin{tabular}{c|c|l|c|c}
\Xhline{1.55pt}
\textbf{Dataset}                                                                & \textbf{Modalities} & \textbf{Property}                                                      & \textbf{Fusion}                                                                            & \textbf{Distilled to}                                                             \\ \Xhline{1.55pt}
\multirow{2}{*}{Affwild2}                                               & Visual     & \begin{tabular}[l]{@{}l@{}}-Strong\\ -Prevalent\end{tabular}  & \multirow{2}{*}{\begin{tabular}[c]{@{}c@{}}Joint Cross \\ Attention\end{tabular}} & \multirow{2}{*}{\begin{tabular}[l]{@{}c@{}}Visual\\ (WES)\end{tabular}}  \\ \cline{2-3}
                                                                       & Audio      & \begin{tabular}[l]{@{}l@{}}-Weak\\ -Privileged\end{tabular}   &                                                                                   &                                                                          \\ \hline
\multirow{2}{*}{\begin{tabular}[l]{@{}l@{}}Biovid\\  (A)\end{tabular}} & Visual     & \begin{tabular}[l]{@{}l@{}}-Weak\\ -Prevalent\end{tabular}    & \multirow{2}{*}{\begin{tabular}[c]{@{}c@{}}Feature\\ Concatenation\end{tabular}}  & \multirow{2}{*}{\begin{tabular}[l]{@{}l@{}}Visual\\ (SEW)\end{tabular}}  \\ \cline{2-3}
                                                                       & EDA        & \begin{tabular}[l]{@{}l@{}}-Strong\\ -Privileged\end{tabular} &                                                                                   &                                                                          \\ \hline
\multirow{2}{*}{\begin{tabular}[l]{@{}l@{}}Biovid\\ (B)\end{tabular}}  & Visual     & \begin{tabular}[l]{@{}l@{}}-Strong\\ -Prevalent\end{tabular}  & \multirow{2}{*}{\begin{tabular}[l]{@{}l@{}}Multimodal\\ Transformer\end{tabular}} & \multirow{2}{*}{\begin{tabular}[l]{@{}l@{}} - Visual (WES) \\ - EMG (SEW)\end{tabular}} \\ \cline{2-3}
                                                                       & EMG        & \begin{tabular}[l]{@{}l@{}}-Weak\\ -Privileged\end{tabular}   &                                                                                   &                                                                          \\ \Xhline{1.55pt}
\end{tabular}
\caption{Snapshot of the diverse modalities, fusion architectures, and distillation settings: Weaker Enhancing Stronger (WES) and Stronger Enhancing Weaker (SEW).}
\label{tab:summ-mod}
\end{table}

\begin{table}[!h]
\centering 
\begin{tabular}{cc|cc}
\Xhline{1.55pt}
\multicolumn{2}{c|}{\textbf{Method}}                                                                                                     & \multicolumn{2}{c}{\textbf{Performance}} \\  
\multicolumn{2}{c|}{}                                                                                                                    & Biovid (B)             & Affwild2             \\ \Xhline{1.55pt}
\begin{tabular}[c]{@{}c@{}}  Aslam et al. \cite{pkd-aslam} \\ \textit{(IEEE CVPRw '23)}\end{tabular}
                           &  Cosine Similarity     &      75.40              &       \begin{tabular}[c]{@{}l@{}}V: 0.37 \\ A: 0.53 \end{tabular} \\  \hline
 \begin{tabular}[c]{@{}c@{}} Makantasis et al. \cite{pkd-makantasis}  \\  \textit{(IEEE TAC '23)}\end{tabular}                 &       MSE              &      N/A           &         \begin{tabular}[c]{@{}l@{}}V: 0.39 \\ A: 0.53\end{tabular} \\  \hline
 \begin{tabular}[c]{@{}c@{}} Makantasis et al. \cite{pkd-makantasis} \\ \textit{(IEEE TAC '23)}\end{tabular}  &  KL                      &      75.80              &     N/A \\               \Xhline{1.55pt}
  \begin{tabular}[c]{@{}c@{}}PKDOT \cite{aslam2024distilling}   \\  \textit{(IEEE FG '24)}\end{tabular}                                              & \begin{tabular}[c]{@{}c@{}}Optimal Transport\\  Structural KD\end{tabular} &        78.76            &      \begin{tabular}[c]{@{}l@{}}V: 0.43 \\ A: 0.56 \end{tabular} \\                \hline
MT-PKDOT (Ours)                                              & \begin{tabular}[c]{@{}c@{}}Multi-Teacher\\  Structural KD\end{tabular} &        \textbf{79.68}            &      \begin{tabular}[c]{@{}l@{}}\textbf{V: 0.44} \\ \textbf{A: 0.56} \end{tabular}
\\
\Xhline{1.55pt}
\multicolumn{4}{l}{N/A: Not applicable to the problem.
}

\end{tabular}
\caption{Performance of the proposed approach and SOTA PKD methods on Biovid (B) and Affwild2 datasets.}
\label{tab:pkd-sota}
\end{table}

We compare the proposed method with SOTA-privileged DL methods in Table~\ref{tab:pkd-sota}. The proposed method outperforms the SOTA PKD methods. MT-PKDOT significantly outperforms the cosine similarity, MSE, and KL-based point-to-point KD methods. PKDOT also improves over the optimal transport-based structural KD. Such an improvement is because the proposed MT-PKDOT method is primarily focused on the corner cases where the multimodal teacher lapses. Since there is still a significant error rate in the upper-bound (multimodal teacher), the student might also learn negatively from cases where the multimodal teacher produces inaccurate predictions. The proposed method overcomes this by providing additional diverse teacher representations as proxy teachers. It can also be observed in Table~\ref{tab:pkd-sota} that, in the case of the Affwild2 dataset (arousal), where the alignment of teachers is not effective, the proposed method is able to maintain performance by falling back to the multimodal teacher.    

\begin{table}[h]
\centering
\setlength{\tabcolsep}{0.45em}
\begin{tabular}{l|c|c|c}
\Xhline{1.55pt}
\textbf{Method}                                                                   & \textbf{Visual Network} & \textbf{Valence} & \textbf{Arousal} \\ \Xhline{1.55pt}
Baseline  \cite{aff-baseline} \textit{CVPRw'20}                                                                      & ResNet-50               & 0.31             & 0.17             \\ \hline
Zhang et al. \cite{Zhang-affwild} \textit{CVPRw'20}                                                                    & SENet-50                & 0.28             & 0.34             \\ \hline
He et al.  \cite{he-affwild} \textit{CVPRw'21}                                                                        & MobileNet               & 0.28             & 0.44             \\ \hline
Nguyen et al.  \cite{nguyen-affwild} \textit{CVPRw'22}                                                                    & RegNet + GRU            & 0.43             & 0.57             \\ \hline
Geesung et al. \cite{geesung} \textit{CVPRw'21}                                                                    & ResNeXt + SENet         & 0.51             & 0.48  
      \\ \hline
Aslam et al. \cite{aslam2024distilling} \textit{IEEE FG'24}                                                                    & I3D        & 0.43             & 0.56 \\ \Xhline{1.55pt}
Visual only (Lower bound) & \multirow{3}{*}{I3D}    & 0.41             & 0.51             \\ \cline{1-1} \cline{3-4} 
\textbf{MT-PKDOT - Student (Ours)}                                                             &                         & \textbf{0.44}          & \textbf{0.56}             \\ \cline{1-1} \cline{3-4} 
\begin{tabular}[c]{@{}c@{}}Multimodal (Visual + Audio)\\ (Upper bound)\end{tabular}                 &                         & 0.67             & 0.59             \\\Xhline{1.55pt}

\end{tabular}
\caption{CCC values for the proposed MT-PKDOT method and SOTA visual-only methods on the validation set of the Affwild2 dataset.}
\label{tab:aff-sota}
\end{table}

Table~\ref{tab:aff-sota} compares the proposed method with SOTA visual-only methods for the Affwild2 dataset. The table shows the lower bound 0.41 valence and 0.51 arousal. The lower bound is the visual-only I3D model without any knowledge transfer. The multimodal teacher, having access to both modalities, achieves 0.67 valence and 0.59 arousal, which serves as the upper bound. The proposed method is able to improve over the lower bound by 3\% percent for valence and 5\% for arousal. It can also be observed that the proposed method improves 1\% over the PKDOT \cite{aslam2024distilling} for valence but maintains the same performance for arousal. This shows that in cases where the teacher alignment is insufficient, the MT-PKDOT can resort back to learning from the multimodal teacher only and maintain performance. This can be attributed to the in-the-wild nature of the Affwild2 dataset, where certain modalities are missing completely, or both the audio and visual modalities are unreliable. There is a significant error rate in the multimodal teacher, which makes the alignment and distillation an even more challenging problem.

 \begin{figure*}[t]
 \captionsetup{justification=centering}
\centering
   \includegraphics[width=1.0\linewidth]{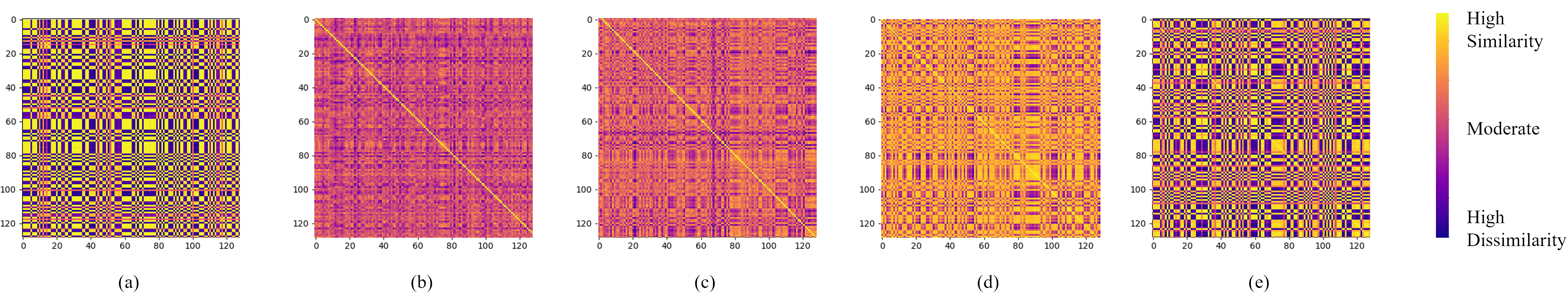}
    \centering
    \caption{Evolution of the student similarity matrix over training epochs. (a) shows the similarity matrix of the pretrained teacher model. (b) through (e) show the student similarity matrix at 0\%, 50\%, 75\%, and 100\% training.   } 

\label{fig:simmat_5}
\end{figure*}

\begin{table}[h]
\centering
\setlength{\tabcolsep}{0.45em}
\begin{tabular}{l|c|cc}
\Xhline{1.55pt}
\multirow{2}{*}{\textbf{Method}}                                                         & \multirow{2}{*}{\textbf{Visual Network}} & \multicolumn{2}{c}{\textbf{Accuracy}}             \\ \cline{3-4} 
                                                                                &                               & \multicolumn{1}{c|}{BL1 vs. PA4} & MC    \\ \Xhline{1.55pt}
Zhi et al.   \cite{zhi-biovid} \textit{ITAIC'19}                                                                  & Sparse LSTM                   & \multicolumn{1}{c|}{61.70}        & 29.70  \\ \hline
Morabit et al.  \cite{morabit-biovida} \textit{ISIVC'22}                                                               & ViT                           & \multicolumn{1}{c|}{72.11}       & NR     \\ \hline
Werner et al. \cite{werner-biovida} \textit{ACIIW'17}                                                                 & FAD + RF                      & \multicolumn{1}{c|}{72.40}        & 30.80  \\ \hline
Patania et al. \cite{patania-biovid_a} \textit{SIGAPP'22}                                                                & GNN                           & \multicolumn{1}{c|}{73.20}        & NR     \\ \hline
Gkikas et al. \cite{gkikas-biovida} \textit{EMBC'23}                                                              & ViT                           & \multicolumn{1}{c|}{73.28}       & 31.52 \\ \hline
 Dragomir et al. \cite{dragomir-biovidA} \textit{EHB'20}                                                               & ResNet-18                     & \multicolumn{1}{c|}{NR}           & 36.60  \\
 \hline
 Aslam et al. \cite{aslam2024distilling} \textit{IEEE FG '24}                                                               & R3D                    & \multicolumn{1}{c|}{74.55}           & 33.65  \\ \Xhline{1.55pt}
Visual only (Lower bound) & \multirow{3}{*}{R3D}          & \multicolumn{1}{c|}{72.10}        & 30.38  \\ \cline{1-1} \cline{3-4} 
\textbf{MT-PKDOT - Student (ours) }                                                       &                               & \multicolumn{1}{c|}{\textbf{75.20}}         & \textbf{33.90}  \\ \cline{1-1} \cline{3-4} 
\begin{tabular}[c]{@{}c@{}}Multimodal (Visual + EDA)\\ (Upper bound)\end{tabular}            &                               & \multicolumn{1}{c|}{83.50}        & 36.40   \\ \Xhline{1.55pt}
\multicolumn{4}{l}{\scriptsize NR: Not Reported.}
\end{tabular}
\caption{Performance comparison of the proposed MT-PKDOT method with SOTA visual-only methods on the Biovid (A) dataset.}
\label{tab:biovida-sota}
\end{table}

Table \ref{tab:biovida-sota} shows the comparison of the proposed method with visual-only SOTA for the Biovid (A) dataset. The lower bound is the visual-only R3D model. The upper bound is the visual + EDA multimodal model. The MT-PKDOT is the student model after distillation. The proposed method improves by 3.1\% over the visual-only baseline. The method also outperforms the SOTA on the Biovid (A) for the two-class problem.

\begin{table}[h]
\centering
\begin{tabular}{l|c|c}
\Xhline{1.55pt}
\textbf{Method}                                                                         & \textbf{Visual Network }      & \textbf{Accuracy} \\ \Xhline{1.55pt}
Kachele et al. \cite{Kchele2015BioVid_mm} \textit{IWMCS'15}                                                                & Handcrafted + RF     & 72.70    \\
\hline
Aslam et al. \cite{aslam2024distilling} \textit{IEEE FG '24}                                                                & Handcrafted + RF     & 78.70    \\ 
\Xhline{1.55pt}

Visual Only (Lower bound) & \multirow{3}{*}{R3D} & 74.10    \\ \cline{1-1} \cline{3-3} 
\textbf{MT-PKDOT - Student  (Ours)}                                                          &                      & \textbf{79.68}    \\ \cline{1-1} \cline{3-3} 
\begin{tabular}[c]{@{}c@{}}Multimodal (Visual + EMG)\\ (Upper bound)\end{tabular}
             &                      & 81.30   \\ \Xhline{1.55pt}
\end{tabular}
\caption{Performance comparison of the proposed MT-PKDOT method with SOTA visual-only methods on the Biovid (B) dataset.}
\label{tab:biovid-sota}
\end{table}

The proposed method is also compared with the SOTA on Biovid (B) in Table~\ref{tab:biovid-sota}. The proposed method improves 5.5\% over the visual-only baseline. The visual-only baseline is 74.10\%, after distillation, the proposed method achieves 79.63\% just by using the visual modality at test time. MT-PKDOT also improves over the structural similarity-based PKD \cite{aslam2024distilling}, showing that the diversity in the teacher module does help improve the performance in the student model. 

It is pertinent to mention here that the aim of this work is not to outperform the SOTA on a particular task. There may be other methods in the literature that are tailored to the task and achieve higher performance in either multimodal or unimodal settings through extensive pertaining and task-specific architecture. The primary goal of this work is to show that with the same backbones, the proposed method is able to improve over the lower bound. More specifically, the introduction of teacher alignment and centroid loss improves performance over the structural similarity-based PKD \cite{aslam2024distilling}.  

To see the effectiveness of the proposed MT-PKDOT method, we also plot and compare the similarity matrices of the teacher and student models. Fig.~\ref{fig:simmat_5} shows the evolution of the student similarity matrix over training epochs.

\begin{figure}[!h]
 \centering
  \includegraphics[width=1\linewidth]{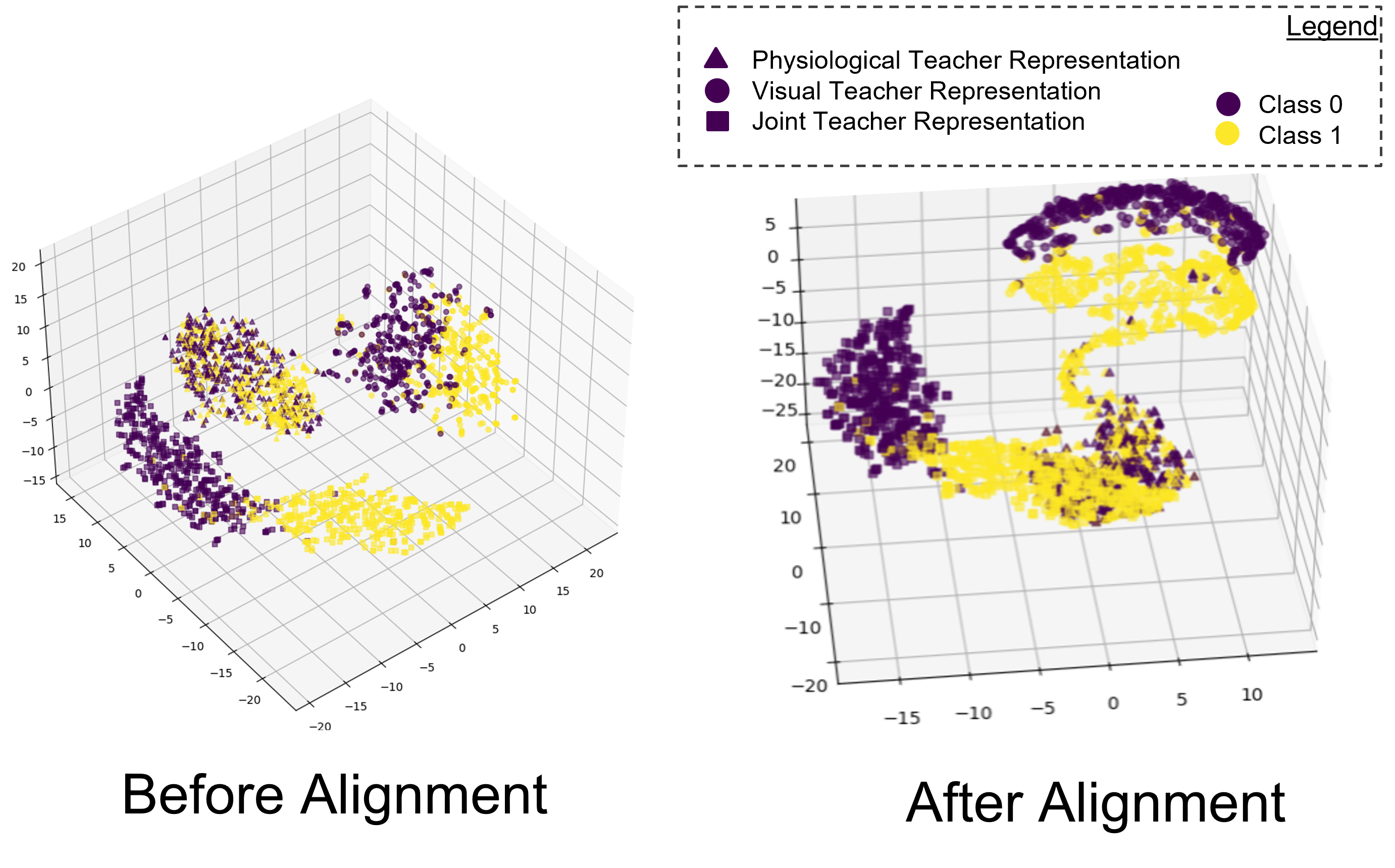}
  \caption{Visualization of the teacher representations before (left) and after (right) alignment on the Biovid (B) dataset}
  \label{fig:teacher-align}
\end{figure}

\begin{figure}[!h]
 \centering
  \includegraphics[width=1\linewidth]{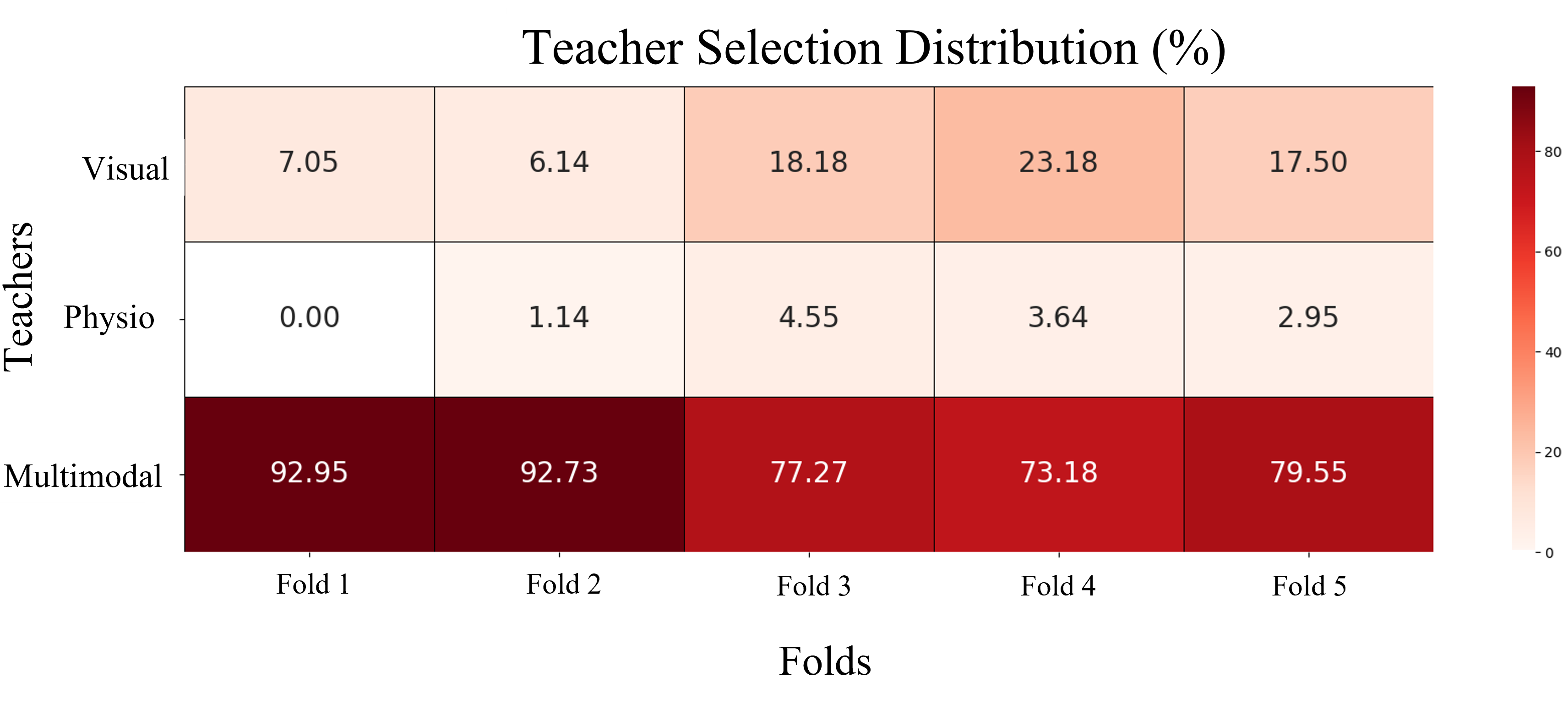}
  \caption{Distribution of teacher selection on the Biovid (B) dataset.}
  \label{fig:per-dis-tab}
\end{figure}

Fig.~\ref{fig:teacher-align} shows the 3D t-SNE plot for the three teacher representations before and after alignment on the Biovid (B) dataset. Before the teachers are aligned with self-distillation, the representations are well-separated discrete clusters because the backbones are also pretrained, which allows the separation of the two classes. After alignment, the three teacher representations are pushed together. The samples belonging to the same class from different teacher networks overlap, showing the alignment of teacher networks while maintaining class separability. Another interesting observation is that the alignment process also increases the discriminating ability of the backbones. This phenomenon is also observed in Fig.~\ref{fig:per-dis-tab}, which shows that the backbone representations also have high discriminatory ability in some instances.

Fig.~\ref{fig:per-dis-tab} shows the distribution of each teacher model selected per batch. This shows that the multimodal teacher model is not always the most accurate teacher, and the MT-PKDOT method selects the aligned visual teacher and the physiological teacher for 14\% and 3\% on average across 5 folds. The proposed method can improve over the single-teacher architecture \cite{aslam2024distilling} because there was no mechanism for the mitigation of negative transfer, and the student was learning from all instances even where the multimodal teacher was not the most accurate.

\subsection{Ablations}
\subsubsection{Batch Size and Anchor Selection}
\begin{figure}[!h]
 \centering
  \includegraphics[width=0.8\linewidth]{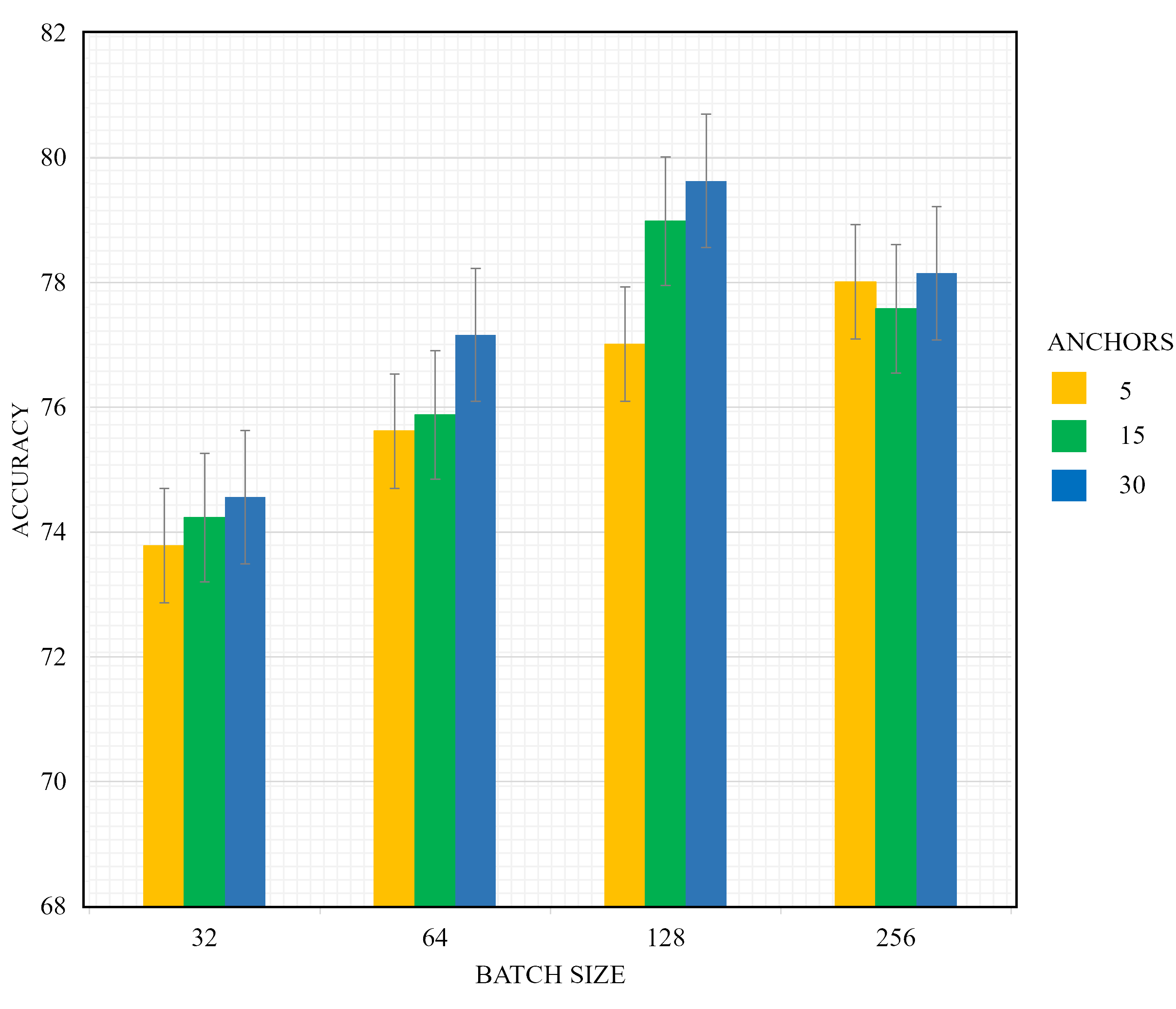}
  \caption{Evolution of accuracy w.r.t increasing batch size and number of anchors on the Biovid (B) dataset.}
  \label{fig:batch-anchor}
\end{figure}

\
Batch size is a particularly crucial hyperparameter in the proposed method since it determines the number of samples that are included in the similarity matrix. Since relational information between the samples in a batch is calculated, the number of samples included at each optimization step can have a substantial effect on the performance. We run the proposed method with different batch sizes. Fig.~\ref{fig:batch-anchor} shows the evolution of the results with respect to increasing batch sizes and anchors. The best results for Biovid (B) are achieved using batch size 128 and 30 anchors.

\subsubsection{Centroid Loss}
One of this paper's contributions is explicitly aligning the centroids of the two representations by introducing an additional constraint to the loss function. We run the proposed method with and without the centroid loss to see its effectiveness. As shown in Tab. \ref{tab:ablation-fr} The student model trained with centroid loss achieves 79.68\%, whereas the student model trained without the centroid loss achieves 79.03\%. This shows the effectiveness of this additional regularization through the alignment of centroids in the teacher and student spaces. 

\begin{table}[h]
  \centering
\begin{tabular}{l|c}
\hline
\textbf{Method} & \textbf{Acc}  \\ \hline
MT-PKDOT w/o $\mathcal{L}_{centroid}$      & 79.03                 \\  
MT-PKDOT w/ $\mathcal{L}_{centroid}$ &   79.68 \\ \hline
\end{tabular}
\caption{Ablation on the centroid loss constraint. The experiment was performed on the Biovid (B) dataset.}
\label{tab:ablation-fr}
\end{table}

\subsection{Distillation to Physiological Modality:}
To further analyze the effectiveness of the proposed method, we also distill to physiological modality. The student model is the same as the physio backbone in the multimodal teacher. The T-Net is trained in reverse to hallucinate the visual modality features. The upper bound is the multimodal teacher model with 81.30\%, and the lower bound is the physio-only model, achieving 63.85\% on the Biovid (B) dataset. After distillation, the MT-PKDOT student is able to improve 2.2\% over the physio-only baseline. Table \ref{tab:biovid-sota-phy} shows the performance of the proposed MT-PKDOT method on the physiological modality.

\begin{table}[h]
\centering
\begin{tabular}{c|c|c}
\Xhline{1.55pt}
\textbf{Method}                                                                         & \textbf{ Network }      & \textbf{Accuracy} \\ \Xhline{1.55pt}

Aslam et al. \cite{aslam2024distilling} \textit{IEEE FG '24}                                                                & ResNet-18 & 65.70    \\ 
\Xhline{1.55pt}

Physio Only (Lower bound) & \multirow{3}{*}{ResNet-18} & 63.85    \\ \cline{1-1} \cline{3-3} 
\textbf{MT-PKDOT - Student  (Ours)}                                                          &                      & \textbf{66.10}    \\ \cline{1-1} \cline{3-3} 
\begin{tabular}[c]{@{}c@{}}Multimodal (Visual + EMG)\\ (Upper bound)\end{tabular}
             &                      & 81.30   \\ \Xhline{1.55pt}
\end{tabular}

\caption{Performance comparison of the proposed MT-PKDOT method with SOTA visual-only methods on the Biovid (B) dataset}
\label{tab:biovid-sota-phy}
\end{table}

\section{Conclusion}
This paper presents a novel multi-teacher privileged knowledge distillation method based on optimal transport that distills the structural information from the aligned teacher representations.  Experiments show that the alignment of diverse modality-specific teachers with MT-PKDOT can enhance student performance. The proposed method is validated on two main MER problems: the valence/arousal estimation regression problem on the Affwild2 dataset and the pain classification on the Biovid (A and B) dataset. To exhibit the generalization ability of our proposed MT-PKDOT method, we use a variety of modalities and combine them using three different fusion architectures to obtain the joint representations. The proposed method can outperform the visual-only baseline as well as SOTA PKD methods for all three teacher architectures, showing that the proposed method is modality-agnostic and model-agnostic. Further, we distill in both SEW and WES settings. The MT-PKDOT method performs better in the WES settings because the stronger modality is still available to the student model. 

A limitation of this work is that the modality alignment does not incorporate any explicit criterion to gauge diversity. This would align modality-specific representations with the joint representation to a reasonable extent. In the future, we intend to develop methods that can employ more modalities in the teacher and student space. More sophisticated attention-based methods can also be applied in the teacher alignment step to achieve better alignment.

\label{sec:conc}



%



\section*{Acknowledgment}

This research endeavor was partially supported by the Natural Sciences and Engineering Research Council of Canada (NSERC), Fonds de recherche du Québec – Santé (FRQS), Canada Foundation for Innovation (CFI), and the Digital Research Alliance of Canada.

\ifCLASSOPTIONcaptionsoff
  \newpage
\fi



\bibliographystyle{IEEEtran}
\bibliography{IEEEtran/bib/mtpkdot}
%



%
\vspace*{-1.1cm}
\begin{IEEEbiography}
[{\includegraphics[width=1in,height=1.25in,clip]{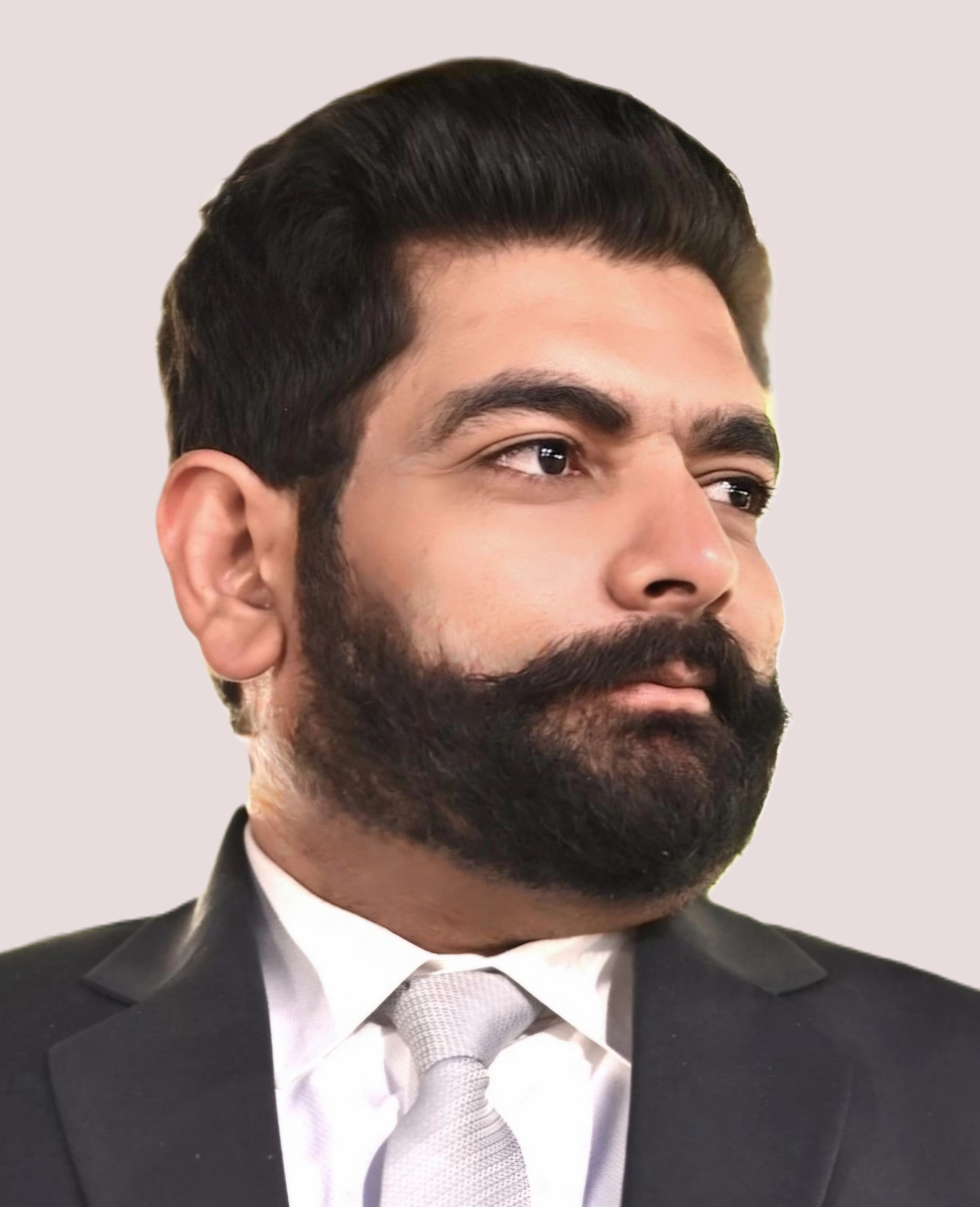}}]{Muhammad Haseeb Aslam} received the Bachelors in Computer Engineering degree (cum laude) and Masters in Data Science (Summa Cum Laude) from Bahria University Islamabad. He is currently pursuing his Doctoral Degree in Engineering from Ecole de Technologie Superiere (ETS) Montreal, Canada. He has mentored various teams and represented Pakistan at multiple prestigious international robotics competitions. His research interests include deep learning, multimodal learning, medical signal processing, and affective computing. He has received international awards, including Richard E. Merwin award for his volunteer work in IEEE. He was also listed amongst rising talents of Pakistan in TechJuice as its class of 25 under 25 young technology leaders.

\end{IEEEbiography}
\vspace*{-1.1cm}
\begin{IEEEbiography}
[{\includegraphics[width=1in,height=1.25in]{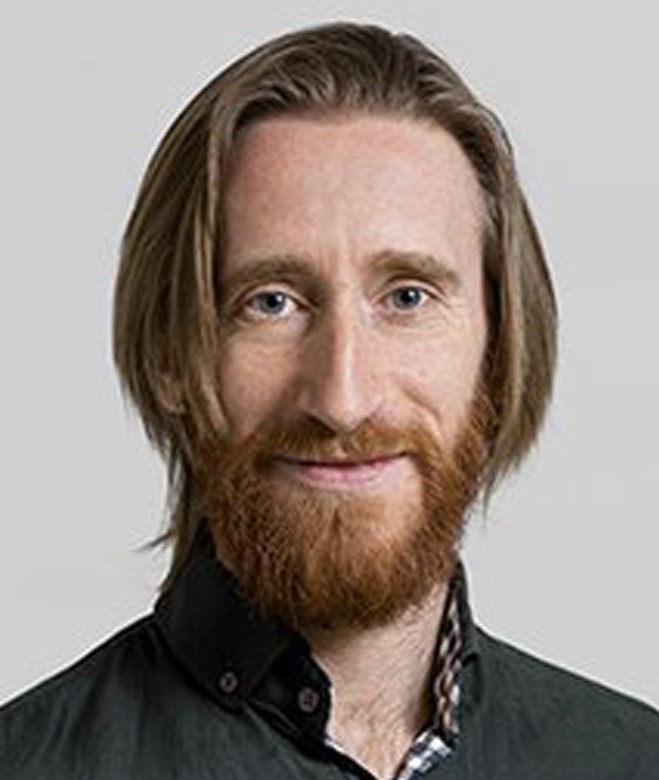}}]{Marco Pedersoli} (Member, IEEE) received the PhD degree in
computer science from the Autonomous University of Barcelona and the Computer Vision Center, in 2012. He was research fellow in computer
vision and machine learning at KU Leuven with
professor Tinne Tuytelaars and later at INRIA
Grenoble with Doctors Jakob Verbeek and Cordelia Schmid. Since 2017, he is associate professor at ETS Montreal, where he is part of LIVIA,
the laboratory of vision and artificial intelligence.
He has authored more than 30 articles in international conferences and computer vision journals.

\end{IEEEbiography}
\vspace*{-1.1cm}
\begin{IEEEbiography}
[{\includegraphics[width=1in,height=1.25in]{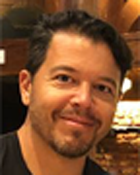}}]{Alessandro Lameiras Koerich} (Member, IEEE) received the Ph.D. degree in engineering from the École de Technologie Supérieure (ÉTS), in 2002. From 1997 to 1998, he was a Lecturer at the Technological Federal University of Paraná, Brazil. From 1998 to 2002, he was a visiting scientist at the CENPARMI, Montréal, Canada. From 2003 to 2015, he was with the Pontifical Catholic University of Paraná, Brazil, where he became a Professor in 2010. From 2006 to 2008, he was the chair of graduate studies in CS. From 2009 to 2012, he was a visiting researcher at INESC-Porto, Portugal. From 2009 to 2015, he was an associate professor at the EE department at the Federal University of Paraná. He has also served as a Fulbright visiting professor with the EE department at Columbia University, USA, in 2013 and as a visiting professor of the EE department at Johns Hopkins University, USA, in 2024. He is currently a professor at the Department of Software and IT Engineering, ÉTS, University of Québec, Montreal. His current research interests include multimodal and trustworthy machine learning, and affective computing. He is an associate editor of the IEEE Transactions on Affective Computing, Pattern Recognition, and Expert Systems with Applications. He has served as the general chair of the 14th International Society for Music Information Retrieval Conference, which was held in Curitiba, Brazil, in 2013. In 2004, he was nominated as an IEEE CS Latin America Distinguished Speaker.

\end{IEEEbiography}
\vspace*{-1.1cm}

\begin{IEEEbiography}[{\includegraphics[width=1in,height=1.25in,clip]{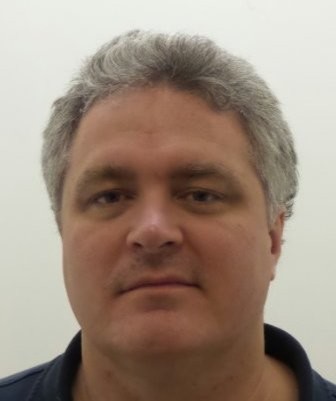}}]{Eric Granger}
(Member, IEEE) received the
Ph.D. degree in EE from École Polytechnique de
Montréal in 2001. He was a Defense Scientist
with DRDC, Ottawa, from 1999 to 2001, and in
Research and Development with Mitel Networks
from 2001 to 2004. He joined the Department
of Systems Engineering, École de technologie
supérieure, Montreal, Canada, in 2004, where he is
currently a Full Professor and the Director of LIVIA,
a research laboratory focused on computer vision
and artificial intelligence. He is the FRQS Co-Chair
in AI and Health, and the ETS Industrial Research Co-Chair on embedded neural networks for intelligent connected buildings (Distech Controls
Inc.). His research interests include pattern recognition, machine learning,
and computer vision, with applications in affective computing, biometrics,
face recognition, medical image analysis, and video surveillance.
\end{IEEEbiography}




\end{document}